\documentclass[11pt,a4paper]{article}
\usepackage[hyperref]{emnlp2020}
\usepackage{times}
\usepackage{latexsym}

\usepackage{url}
\usepackage{algorithm}
\usepackage{algpseudocode}
\usepackage{amsmath}
\usepackage{graphicx}
\usepackage{booktabs}
\usepackage{subcaption}
\usepackage[T1]{fontenc}
\usepackage[utf8]{inputenc}

\usepackage{microtype}
\aclfinalcopy 
\def\aclpaperid{3051} 

\newcommand{\Ym}{Y_{{mask}}}
\newcommand{\Yo}{Y_{{obs}}}

\title{Inference Strategies for Machine Translation with Conditional Masking}

\author{Julia Kreutzer \and
  George Foster \and
  Colin Cherry\\
  Google Research \\
  \texttt{\{jkreutzer,fosterg,colincherry\}@google.com}
  }

\date{}

\begin{document}
\maketitle
\begin{abstract}
  Conditional masked language model (CMLM) training has proven successful for non-autoregressive and semi-autoregressive sequence generation tasks, such as machine translation. Given a trained CMLM, however, it is not clear what the best inference strategy is. We formulate masked inference as a factorization of conditional probabilities of partial sequences, show that this does not harm performance, and investigate a number of simple heuristics motivated by this perspective. We identify a thresholding strategy that has advantages over the standard ``mask-predict'' algorithm, and provide analyses of its behavior on machine translation tasks.  
\end{abstract}

\section{Introduction}
The widely successful masked language modeling paradigm popularized by BERT~\cite{devlin-etal-2019-bert} has recently been adapted to conditional masked language model (CMLM) training for semi-autoregressive sequence generation
\cite{ghazvininejadETAL:19}, where model predictions are conditioned on the complete input sequence and the observed (non-masked) portion of the output sequence.
The CMLM's simplicity and its clear links to the very active field of linguistic representation learning are advantages over its semi-autoregressive competitors, such as iterative refinement of token sequences~\cite{leeETAL:18}, refinement of non-linguistic intermediate representations~\cite{kaiser2018fast,shu2020latent} and learning to predict parallel edit operations~\cite{stern2019insertion,gu2019levenshtein}.

It is not obvious how to best perform inference with the CMLM. Starting from a partially-observed output sequence, the optimal choice to complete it within a single step would be to generate the most likely token at each unobserved (masked) position independently.
However, it is less clear how to progress from an initial, completely masked sequence to a final hypothesis semi-autoregressively over a number of steps, with each successive step unmasking new context for the next. This requires not only ordering the tokens for generation, but also making decisions about how many tokens to simultaneously predict in each step. 

\newcite{ghazvininejadETAL:19} propose the {\em mask-predict} algorithm, which iteratively generates fresh model predictions for all masked positions, and then unmasks a predefined number of the most likely predictions. Given a fixed number of iterations, a decaying schedule determines how many predictions to unmask in each iteration. 
Each successive iteration provides mode-breaking~\cite{guETAL:18} context for the next.
By fixing the number of iterations, this approach allows for constant-time semi-autoregressive decoding.

The fixed-iteration strategy is very practical and has yielded empirical success in a range of machine translation experiments, but there is no guarantee that it is optimal.
The tokens to be unmasked on a given iteration are all predicted independently, and therefore might contain repeated words, or words with low model confidence. These issues can be mitigated by later re-masking a token to repair it~\cite{ghazvininejadETAL:19} or by adapting the model to incorrect contexts~\cite{Ghazvininejad2020SemiAutoregressiveTI}.

We instead adopt a fully probabilistic view of the masked prediction sequence, which we enable by simply disallowing the re-masking of previously unmasked tokens. This view guides us to a heuristic inference schedule that selects sets of unmasked tokens according to a threshold on the product of their conditionally independent model probabilities. This heuristic naturally slows down in the situations mentioned above, and speeds up in the presence of high confidence, which allows us to achieve favorable quality-to-speed trade-offs. 
We focus on strengthening the CMLM inference (Section~\ref{sec:cmlm_inference}) while leaving its training algorithm unchanged (Section~\ref{sec:cmlm_training}), and maintaining much of the structure of the original inference strategy. For our experiments on machine translation (Section~\ref{sec:experiments}), we compare inference heuristics in terms of their quality-speed trade-offs. We analyze the development of quality over iterations, and the influence of sentence length. With examples of unmasking schedules we furthermore illustrate the role of mode breaking through choosing the right contexts.

\section{CMLM Model and Training}
\label{sec:cmlm_training}
The CMLM is a model for $p(\Ym|\Yo, X)$, the probability of masked tokens $\Ym$ given a partially observed output sequence $\Yo$ and an input sequence $X$. 
$\Ym$ and $\Yo$ are sets of tokens at specified positions that together form a complete output sequence $Y$: $\Ym = Y\setminus \Yo$.
The model is implicitly conditioned on output sequence length $N = |Y|$,
and the tokens in $\Ym$ are conditionally independent:
$p(\Ym|\Yo, X) = \prod_{y_i\in \Ym} p(y_i|\Yo,X,N)$.
During training, masks are placed randomly: First, the mask size $S \in \{1,\ldots, N\}$ is sampled from a uniform distribution, then $S$ positions are randomly chosen to define the subsets $\Yo$ and $\Ym$.
Cross-entropy loss is incurred via $p(y_i|\Yo,X)$ for each $y_i\in \Ym$. An additional classifier on top of encoder representations is trained to predict the output length $N$.

\section{CMLM Inference}\label{sec:cmlm_inference}

Inference starts with a context of only MASK tokens. Until a stop condition is met, decoder predictions iteratively replace a subset of these in selected positions (``unmasking''). 
With a single iteration, inference is non-autoregressive; when the number of iterations $T$ is less than the sentence length $N$ it is semi-autoregressive; and when $N=T$ it is fully autoregressive. Due to the use of a uniform distribution over reference contexts, training is agnostic to these different regimes. 

In general, we seek to minimize $T$ without trading off too much quality. The challenge in doing so is to identify the subset of predictions that are most likely to provide suitable conditioning context for future iterations \cite{mansimov2019generalized}. Structural or linguistic dependencies in the output may also play an important role for resolving linguistic ambiguities~\citep{martins-kreutzer-2017-learning}. For example, in German it might be harder to first generate the determiner before knowing the grammatical gender of the head word (see examples in Figure~\ref{fig:examples-main}).

The length predictor first predicts $b$ different lengths, then one hypothesis is decoded for each length independently using the iterative process just outlined. The hypothesis with the highest length-normalized model score is selected as the output. We refer to $b$ as the \textit{length beam} in the following.

\subsection{Update Strategies}
\label{sec:updates}
The CMLM can make predictions at all positions,  whether they correspond to masked input or not. This lends itself to various strategies for choosing how to update current predictions and masks:\footnote{In all cases we assume predictions to be the most likely words at each position, and scores to be the corresponding probabilities.}
\begin{itemize}
\item \textit{update-all}: update tokens and scores at all positions, no constraint on new mask\footnote{This corresponds to a masked version of iterative refinement~\cite{leeETAL:18}.}
\item \textit{update-masked}: update tokens at masked positions only, no constraint on new mask\footnote{This is the strategy used by \newcite{ghazvininejadETAL:19}.}
\item \textit{update-masked-sub}: update tokens at masked positions only, new mask must be a subset of the current one
\end{itemize}

\begin{figure}
\centering
\begin{tabular}{l|l|l|l}
$t$ & $M^{(t)}$ & $Y^{(t)}$ & $p(Y^{(t)}|Y^{(<t)},M^{(\leq t)}, X)$ \\
\hline
0 & \{1,2,3\} & \{\} & -- \\
1 & \{\} & \{a,b,c\} & p(a|X) p(b|X) p(c|X) \\
\hline
0 & \{1,2,3\} & \{\} & -- \\
1 & \{2,3\} & \{a\} & p(a|X) \\
2 & \{3\} & \{b\} & p(b|a,X) \\
3 & \{\} & \{c\} & p(c|a,b,X) \\
\hline
0 & \{1,2,3\} & \{\} & -- \\
1 & \{2\} & \{a,c\} & p(a|X) p(c|X) \\
2 & \{\} & \{b\} & p(b|a,c,X) \\
\end{tabular}
\caption{Computations for $p(Y, M|X)$, where $Y=\{a, b, c\}$, for 
various masking sequences $M$. The first sequence is fully non-autoregressive, 
and the second is the standard left-to-right autoregressive factorization.
}
\label{fig:masking-example}
\end{figure}

In this paper we focus on the {\em update-masked-sub strategy}. It is empirically competitive (Section~\ref{sec:experiments:update_strategies}), and interesting because it corresponds to a valid probabilistic factorization of the target sequence, governed by a latent variable $M = M^{(0)}\ldots M^{(T)}$ which represents 
the sequence of masking decisions:
\begin{multline} \label{eqn:latent-m}
p(Y,M|X) = \prod_{t=1}^T p(Y^{(t)}|Y^{(<t)},M^{(\leq t)},X) \times \\
 p(M^{(t)}|Y^{(<t)},M^{(<t)},X),
\end{multline}
where $M^{(0)}=\{1,\ldots,N\}$, $M^{(T)}=\{\}$, $M^{(t)}\subset M^{(t-1)}$, and  $Y^{(t)}$ is the set of tokens {\em unmasked} on the $t$\/th iteration.  Figure~\ref{fig:masking-example} illustrates this computation for various choices of $M$.\footnote{Note that a probabilistic interpretation enables an unconstrained search for the most probable output, or for the unmasking sequence that assigns highest probability to a reference output, options we do not pursue in this paper.}

The class of inference strategies we explore can thus be seen as \textit{greedy search} for the mostly likely factorization, subject to a constraint on the number of iterations: at each iteration, we choose a subset of tokens to add to the current hypothesis, balancing high model probabilities with the risk of making an error and degrading future predictions. Because tokens are predicted independently, the risk of an error grows with the size of the subset.

\subsection{Unmasking Heuristics}

Under the \textit{update-masked-sub} constraint, the role of greedy inference heuristics is to choose which positions to unmask, given a full set of predictions for all currently-masked positions. The {\em mask-predict} strategy of \newcite{ghazvininejadETAL:19} chooses the $\lceil N/T\rceil$ highest-probability tokens, in order to finish in a constant $T$ iterations, regardless of $N$. 
This generates more tokens per iteration for long sentences, which may not be ideal for sentences with complex structure. To measure its effect, we propose a variant that unmasks a constant $K$ tokens per iteration, in order to achieve approximately $K$-fold speedup over autoregressive performance, independent of hypothesis length. 

Unmasking highest-ranked tokens according to probability is reasonable, but it ignores the magnitude of the probabilities, creating the potential for selecting tokens in which the model has low confidence, and vice versa. To address this, we design several simple thresholding strategies that vary the number of tokens per iteration, ideally generating more when the conditioning context licences many confident predictions, and fewer otherwise. 
\begin{enumerate}
    \item The most straightforward strategy, {\em thresh}, unmasks all tokens with probabilities greater than a given threshold $\tau$.
    \item The {\em comb-thresh} strategy unmasks the largest set of highest-ranked tokens $Y$ whose joint probability $p(Y) > \tau$. 
    \item Finally, in order to account for lower-ranked predictions, the {\em fcomb-thresh} strategy unmasks the largest set $Y$ for which $p(Y) * (1 - p(\bar{Y})) > \tau$, where $Y$ consists of the highest-ranked tokens, and $\bar{Y}$ is its complement.
\end{enumerate} 
All threshold strategies unmask the single highest-ranked token in contexts where the threshold criterion is not met.

\section{Experiments}\label{sec:experiments}

Our CMLM is implemented with a base Transformer~\citep{vaswaniETAL:17} built on a TensorFlow implementation of \cite{ghazvininejadETAL:19}. 
The input to the decoder is $\Yo$, with MASK tokens at masked positions, and the output is $\Ym$, predictions for all masked positions without future attention masking. 
We use data from WMT14 en$\leftrightarrow$de~\citep{bojar-EtAl:2014:W14-33} 
and WMT17 zh$\leftrightarrow$en~\citep{bojar-EtAl:2017:WMT1} with a sentence piece vocabulary of 32k, focusing mainly on en$\rightarrow$de, and providing results for all pairs
in appendix~\ref{app:experiments}.
The CMLM is trained on distilled training data from an autoregressive Transformer and initialized with its parameters. 

\subsection{Update Strategies}
\label{sec:experiments:update_strategies}
\begin{figure}[h]
    \centering
    \includegraphics[width=\columnwidth]{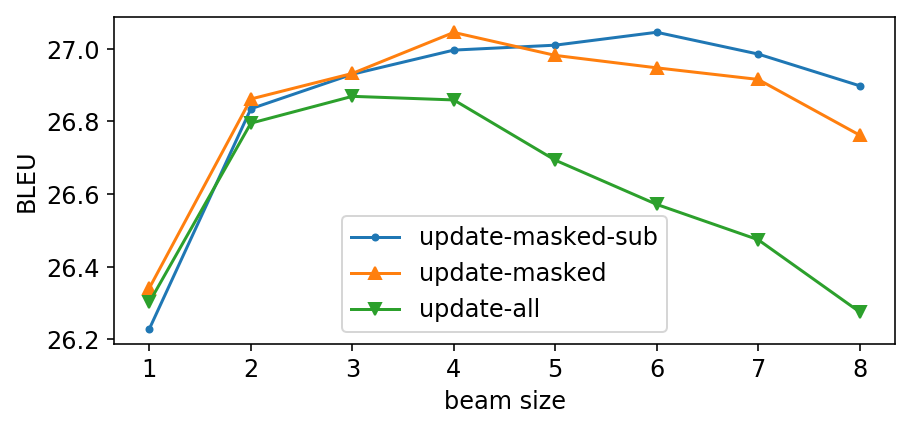}
    \caption{Performance of update strategies}
    \label{fig:new-beams}
\end{figure}

Figure~\ref{fig:new-beams} shows the performance of the update strategies described in section~\ref{sec:updates} versus length beam $b$.
All strategies use the mask-predict heuristic with a fixed 10-iteration limit. 
As beam size increases past 2, the update-masked strategies increasingly dominate, indicating that their scores are more reliable for choosing among length hypotheses. There is no significant difference between the two variants of update-masked. This suggests that our probabilistic factorization constraint (\emph{update-masked-sub}) does not hurt in practice.

\subsection{Heuristics}

\begin{figure}[t]
    \centering
    \includegraphics[width=\columnwidth]{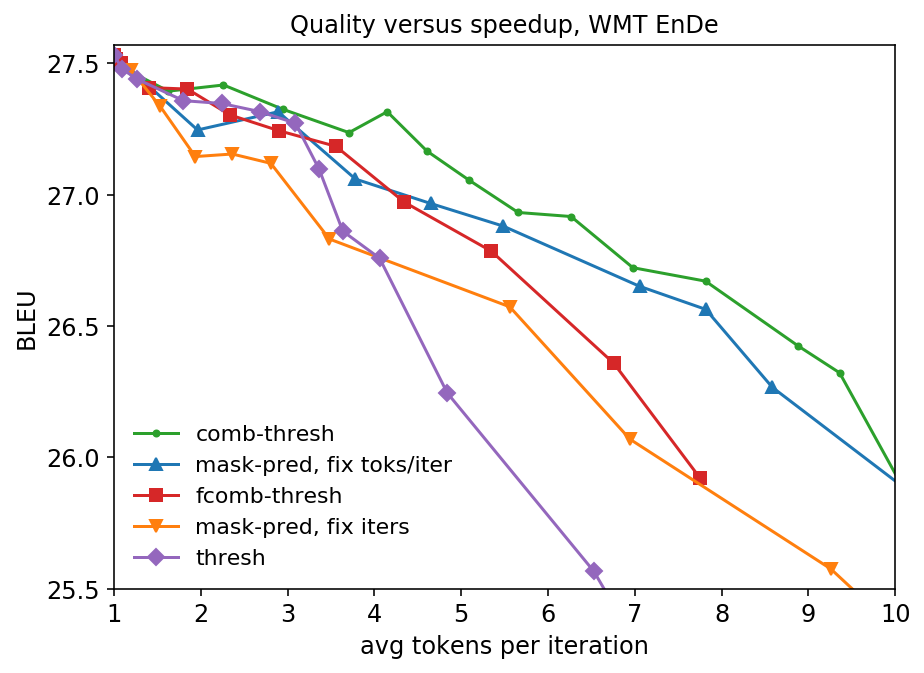}
    \caption{Heuristics with length-beam=5. The top border of the graph represents the performance of an autoregressive Transformer with beam=5. Different points in the graph correpond to different hyperparameter settings, varying the hyperparameter which controls the speed for each heuristic ($T$ for fixed-iteration \emph{mask-predict}, $K$ for variable-iteration \emph{mask-predict}, and $\tau$ for thresholding heuristics).}
    \label{fig:heuristics2}
\end{figure}
To compare the speed-quality trade-off of different heuristics on an equal footing, we vary the values of the hyper-parameter that controls speedup: $T$ for fixed-iteration mask-predict, $K$ for variable-iteration mask-predict, and $\tau$ for thresholding strategies. In each case, we measure the resulting speedup as the total number of tokens in the test set divided by the total number of iterations required for all sentences,\footnote{This is theoretical speedup, and we make no claims that it can be attained in practice, an objective that would likely require significant engineering effort.} and corpus BLEU on the output of the last iteration.

\begin{figure}[t!]
    \centering
    \includegraphics[width=\columnwidth]{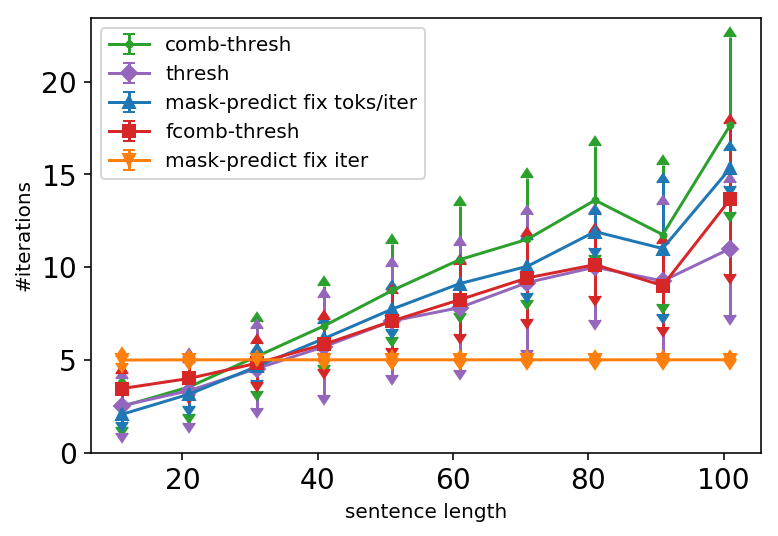}
    \caption{Number of iterations on the test set in relation to (oracle) sentence length (tokens) when generating on average ca. 5 tokens per iteration. Error bars indicate the standard deviation.}
    \label{fig:iterationnumbers}
\end{figure}

\begin{figure}[t!]
    \centering
    \includegraphics[width=\columnwidth]{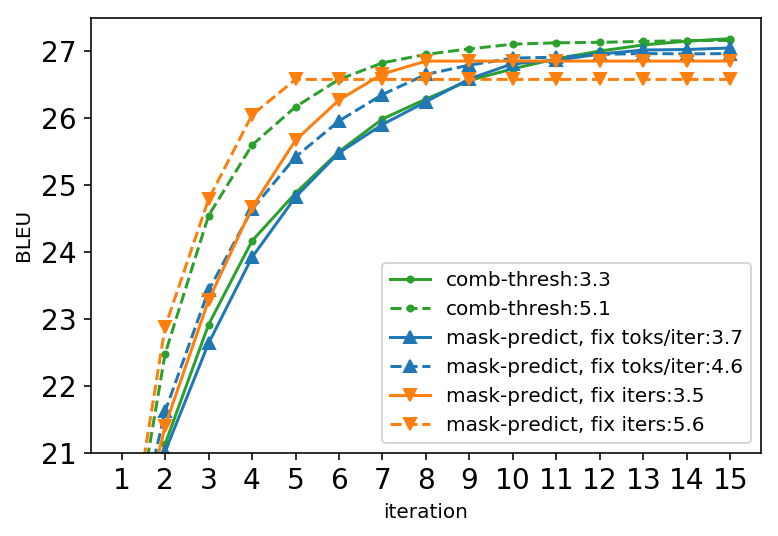}
    \caption{Development of BLEU over iterations comparing heuristics for two different generation speeds. The generation speed is expressed in average number of tokens per iteration, e.g. \texttt{comb-thresh:3.5} stands for the \emph{comb-thresh} heuristic with a threshold value set so that 3.5 tokens are generated per iteration on average.}
    \label{fig:bleu-iters}
\end{figure}

\begin{figure*}[t]
    \centering
     \includegraphics[width=\textwidth]{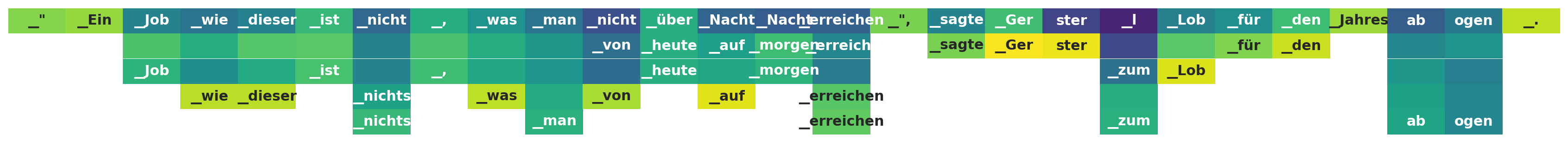}
    \includegraphics[width=\textwidth]{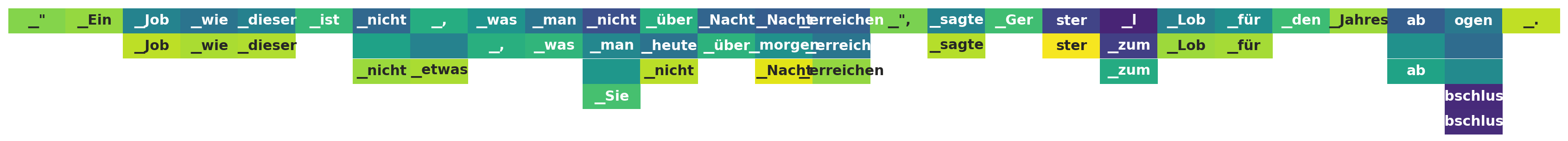}
    \caption{Unmasking over iterations for source sentence \textit{"A job like this is not something you achieve overnight," said Gerster in praise of the annual financial statement.} with reference \textit{"So ein Werk schüttelt man nicht einfach aus dem Ärmel", lobte Gerster mit Blick auf die Jahresrechnung.} for \emph{mask-predict} with $T=5$ above and \emph{comb-thresh} with $\tau=0.1$ below. Predicted tokens are shown when they differ from the previous iteration. Their background color indicates the model score, with yellow indicating high scores, and dark blue low scores.}
    \label{fig:examples-main}
\end{figure*}

Figure~\ref{fig:heuristics2} compares heuristics using 5 length candidates.
First of all, fixed-$K$ {\em mask-predict} beats fixed-$T$ by a substantial margin (especially at higher speeds), indicating that it is worth allocating more iterations for longer sentences.  Second, 
the {\em comb-thresh} strategy has a small but consistent advantage over fixed-$K$ {\em mask-predict} across all speeds. This strategy exhibits a roughly 4x gain while sacrificing less than $0.3$ BLEU relative to the equivalent autoregressive Transformer ($27.6$ BLEU). 

Both {\em thresh} and {\em fcomb-thresh} underperform. 
Despite their superficial similarity to {\em comb-thresh}, they perform much worse; this holds for other language pairs as well (Figure~\ref{fig:all-langs} in Appendix~\ref{app:experiments}). For {\em thresh}, the poor performance as speedup increases reflects many relatively low-probability tokens exceeding lower thresholds, a condition that is penalized by all other heuristics, which take rank into account. For {\em fcomb-thresh} the effect is more subtle; we believe that it is due to the probabilities of lower-ranked tokens having worse calibration, leading to less reliable unmasking decisions. 

A practical impediment to a thresholding strategy is that it does not provide direct control over desired speedup: this must be identified by tuning $\tau$ appropriately on a development set. However, we found that dev and test speedups were  well correlated across speedups ranging from 1 to 11, with the largest absolute error being 0.8 (11.1 speedup on dev versus 10.3 on test), and the average error being 0.3.

\subsection{Analysis}\label{sec:analysis}
Having freed the heuristics from a globally imposed iteration limit for constant-time decoding as in the original mask-predict inference heuristic, we observed better quality-speed trade-offs in the above discussed results. Intuitively, we would expect the heuristics to allocate more iterations for longer sentences and save iterations on shorter sentences.
 Figure~\ref{fig:iterationnumbers} shows how many iterations the models spend on sentences in relation to their length. For a fair comparison, the generation is constrained to oracle output lengths, and we set the hyperparameters such that they result in the same generation speed (5 tokens per iteration on average). We see that flexible-iteration strategies spend fewer iterations on sentences up to a length of around 30 when compared to a fixed-iteration strategy. \textit{comb-thresh} spends on average the largest number of iterations on longer sentences (which pays off in terms of quality, see Figure~\ref{fig:heuristics2}), while \textit{thresh} spends even fewer iterations on longer sentences than the mask-predict model.

The development of BLEU over iterations for comparable generation speeds across heuristics is shown in Figure~\ref{fig:bleu-iters}.\footnote{Each line on this graph is produced by doing inference with a particular hyperparameter setting, and recording BLEU for the greedily predicted tokens after each iteration.} We can see that speedier generation gives a faster initial increase in translation quality over iterations in exchange for slightly lower final quality (dashed vs solid lines). 
{\em Mask-predict} levels off early after reaching its fixed number of iterations, but climbs quickly before that point due to an averaging effect over short sentences.
Fixed-K {\em mask-predict} and {\em comb-thresh} both extract useful work out of each iteration, with {\em comb-thresh} maintaining a slight edge over all iterations, especially at higher generations speeds.

Figure~\ref{fig:examples-main} shows an example for generation strategies under \textit{mask-predict} and \textit{comb-thresh} (see appendix~\ref{app:examples}). They illustrate the workings of iterative decoding and main differences between strategies:
Iterative decoding is crucially needed to resolve subject-verb agreement (e.g. ``man $[\dots]$ erreichen'' (generic ``you'') vs. ``Sie $[\dots]$ erreichen'' (formal ``you'') in ex. 2) and rough sentence structure (e.g. placement of the comma), and offers room for less literal translations (``von heute auf morgen'' (literally ``from today to tomorrow'') rather than ``über Nacht'' (literally ``over night'') in ex. 1).
The two tokens ``Ger'' and ``ster'' (a name) show how the correct conditioning changes model scores in both cases: After the former token is predicted, the probability for the latter increases drastically, since its only valid position in the sentence is there.
While both strategies use the same number of iterations to generate this translation, one can see that it pays off for \textit{comb-thresh} to unmask certain tokens earlier (``ab'', ``Lob''), which allows a valid resolution of neighboring tokens (``bschluss'', ``zum'').\footnote{A typo (``abbschluss'' vs ``abschluss'') is introduced by choosing the ``ab'' sub-word rather than ``a'', likely contributing to the model uncertainty in this area.}

\section{Conclusion}
We investigated inference strategies for machine translation based on CMLM with a focus on the trade-off between generation speed and quality. 
We introduce a perspective which views generation sequences as probabilistic factorizations of the final output sequence, and use it to analyze and extend previous heuristics.
Our new heuristics achieve better speed/quality balance by flexibly adjusting the number of total iterations, and by taking the probabilities of sets of tokens into account.
For future work we would like to explore if their success transfers to other generation tasks with MLMs where inference efficiency is a concern.
\clearpage
\bibliography{references}
\bibliographystyle{acl_natbib}

\appendix

\clearpage
\section{Experiments on different languages}\label{app:experiments}
Pre-processing for all data sets follows the procedure described in ~\citep{vaswaniETAL:17}.

\begin{figure*}
    \centering
    \includegraphics[scale=0.5]{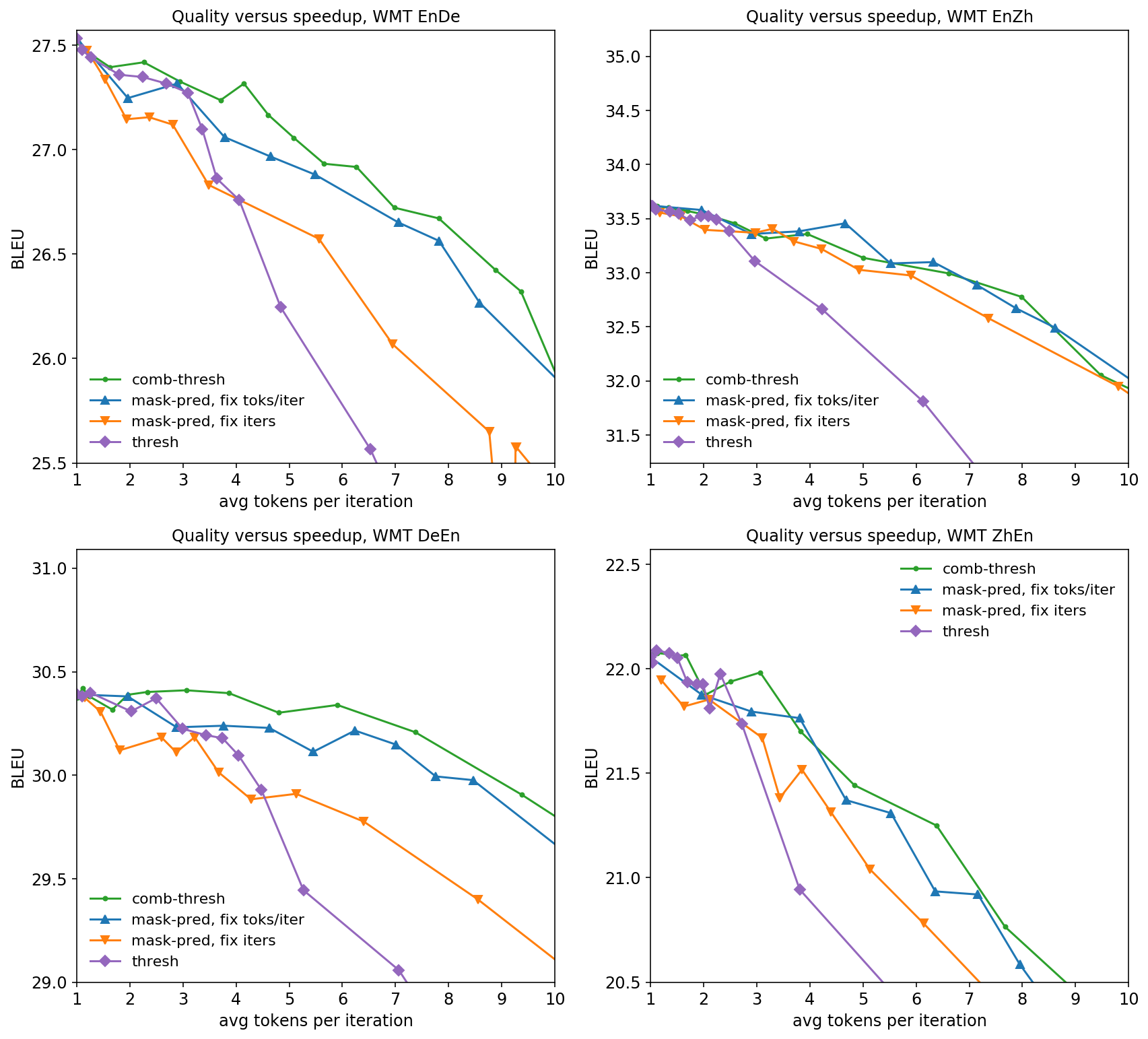}
    \caption{Results on all language pairs, with length-beam 5. The top border of each graph represents performance of the equivalent autoregressive Transformer.}
    \label{fig:all-langs}
\end{figure*}

Figure~\ref{fig:all-langs} shows the results for heuristics on all language pairs. 
As in our main experiments, quality is measured with tokenized BLEU, except for en$\rightarrow$zh, where we use SacreBLEU \cite{Post2018ACF}. In three of the language pairs, we observe a similar pattern to en$\rightarrow$de: {\em comb-thresh} has a slight but consistent advantage over mask-predict, with the fixed tokens/iteration version of {\em mask-predict} doing consistently better than the fixed iteration version. On en$\rightarrow$zh, all three methods perform similarly.

\section{Examples}\label{app:examples}
\begin{figure*}[t]
    \centering
     \centering
     \begin{subfigure}{\textwidth}
         \centering
          \includegraphics[width=\textwidth]{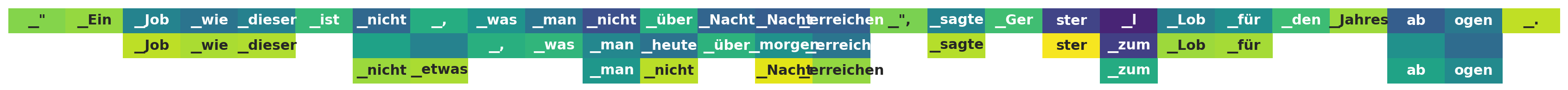}
        \includegraphics[width=\textwidth]{images/mp-2.png}
    \includegraphics[width=\textwidth]{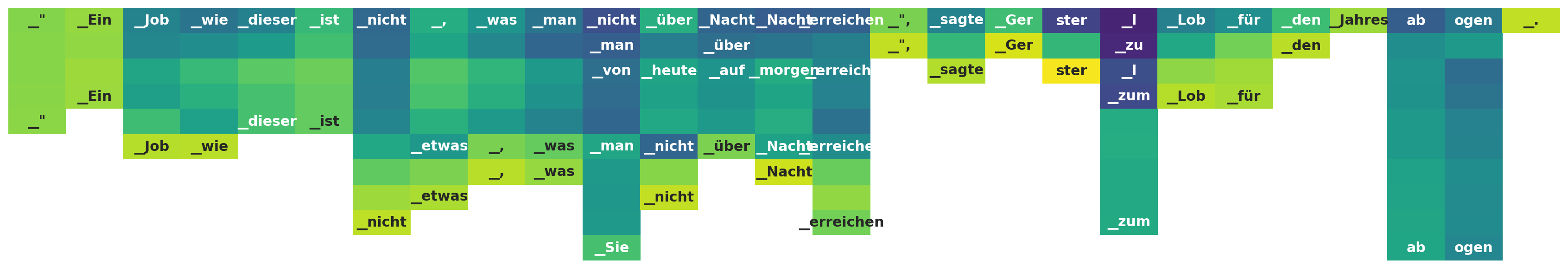}
           \caption{Configurations of \textit{mask predict} with fixed number of iterations: (1) $T=3$, (2) $T=5$, (3) $T=10$. }
         \label{fig:example-mp}
     \end{subfigure}
     \par\bigskip
     \begin{subfigure}{\textwidth}
         \centering
          \includegraphics[width=\textwidth]{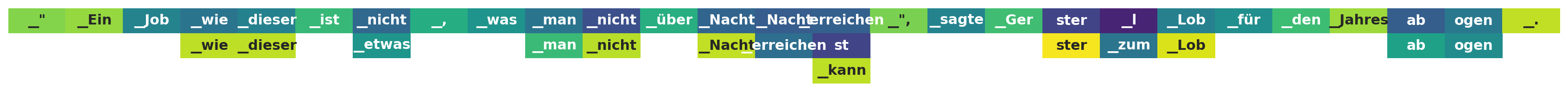}
   \includegraphics[width=\textwidth]{images/comb-thresh-2-big-noy-masked.png} 
   \includegraphics[width=\textwidth]{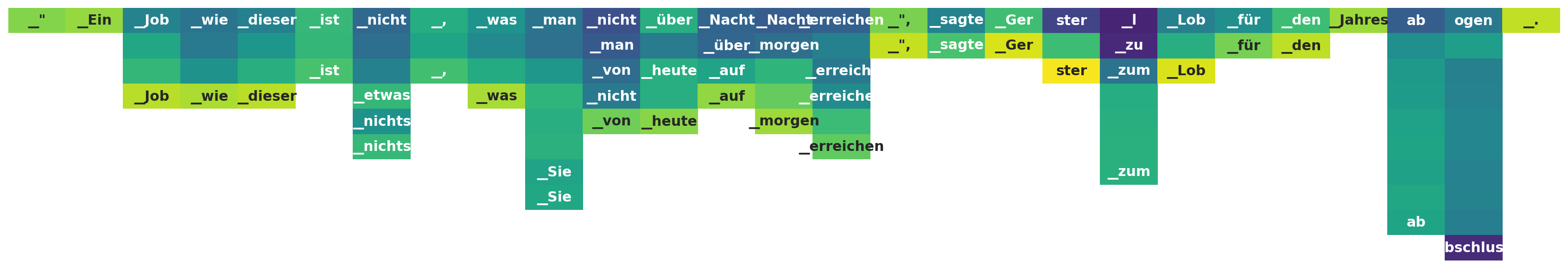}
           \caption{Configurations of \textit{comb-thresh}: (1) $\tau=0.002$, (2) $\tau=0.1$, (3) $\tau=0.4$. }
         \label{fig:example-comb-thresh}
     \end{subfigure}
    \caption{Example unmasking schedules. Predicted tokens are shown when they differ from the previous iteration. Their background color indicates the model score, with yellow indicating high scores, and dark blue low scores.}
    \label{fig:examples-app}
\end{figure*}

Figure~\ref{fig:examples-app} provides more examples for \emph{mask-predict} (Figure~\ref{fig:example-mp}) and \emph{comb-thresh} (Figure~\ref{fig:example-comb-thresh}) heuristics under different hyperparameter settings, complementing the ones displayed in Figure~\ref{fig:examples-main}. The source sentence is \textit{"A job like this is not something you achieve overnight," said Gerster in praise of the annual financial statement.}, and the reference \textit{"So ein Werk schüttelt man nicht einfach aus dem Ärmel", lobte Gerster mit Blick auf die Jahresrechnung.}. 
    Predicted tokens are printed out when they differ from the previous iteration. Their background color indicates the model score, with yellow indicating high scores, and dark blue low scores.

\end{document}